\pdfoutput=1

\documentclass[11pt]{article}
\usepackage{mdframed}

\usepackage{aaai2026}
\usepackage{times}
\usepackage{helvet}
\usepackage{courier}
\usepackage[hyphens]{url} 
\usepackage{graphicx}
\usepackage{natbib} % Этот пакет нужен для цитирования
\usepackage{caption}
\usepackage{graphicx}
\usepackage{amsmath,amssymb}
\usepackage{booktabs,multirow}
\usepackage{enumitem}                  % bullet tuning
\usepackage{natbib}
\usepackage{tabularx}                  % auto-wrapping table cols
\usepackage{pdflscape}                 % optional landscape
\usepackage{times}
\usepackage{latexsym}
\usepackage[table]{xcolor}   % <-- добавлено
\usepackage{colortbl}        % <-- добавлено
\usepackage[T1]{fontenc}
\usepackage{tabularx}      % адаптивная ширина столбцов
\usepackage{booktabs}
\usepackage{float}         % [H] – жёсткая фиксация
\usepackage{placeins}      % \FloatBarrier
\usepackage{caption}       % компактные подписи
\usepackage[utf8]{inputenc}
\nocopyright
\usepackage{microtype}

\usepackage{inconsolata}
\usepackage{graphicx}
\usepackage{hyperref}
\usepackage{fontawesome5}
\hypersetup{
    colorlinks=true,   % <--- эта строка убирает рамки
    allcolors=black,   % <--- эта строка делает текст ссылок черным
    pdftitle={CHECK-MAT},
    pdfpagemode=FullScreen,
}

\title{CHECK-MAT: Checking Hand-Written Mathematical Answers for the Russian Unified State Exam}

\author{
  \textbf{Khrulev Ruslan} \\
  Lomonosov Moscow State University \\
  \href{mailto:ra.khrulev@gmail.com}{\texttt{ra.khrulev@gmail.com}} \\
  \href{https://github.com/Karifannaa/Auto-check-EGE-math}{\faGithub Repository} \quad
  \href{https://huggingface.co/datasets/Karifannaa/EGE_Math_Solutions_Assessment_Benchmark}{\includegraphics[height=1.5em]{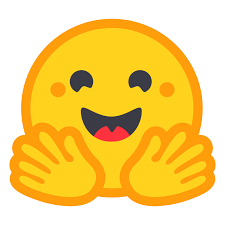} Dataset}
}

\begin{document}
\maketitle
% --------- abstract ------------------------------------------
\begin{abstract}
This paper introduces a novel benchmark, \emph{EGE-Math
Solutions Assessment Benchmark}, for evaluating Vision–Language
Models (VLMs) on their ability to assess hand-written mathematical
solutions. Unlike existing benchmarks that focus on problem solving,
our approach centres on understanding student solutions, identifying
mistakes, and assigning grades according to fixed criteria.  We compile
122 scanned solutions from the Russian Unified State Exam (EGE) together
with official expert grades, and evaluate seven modern VLMs from
Google, OpenAI, Arcee AI, and Alibaba Cloud in three inference modes.
The results reveal current limitations in mathematical reasoning and
human-rubric alignment, opening new research avenues in AI-assisted
assessment. You can find code in \url{https://github.com/Karifannaa/Auto-check-EGE-math}
\end{abstract}

\section{Introduction}
The automated assessment of mathematical solutions, particularly those involving handwritten content and complex reasoning, represents a significant frontier for artificial intelligence. The evolution of this field has moved from deterministic, answer-focused systems to probabilistic models that attempt to evaluate a student's entire problem-solving process. While numerous benchmarks exist to test a model's ability to solve mathematical problems—such as \textit{MATH} \cite{hendrycks2021measuring} or \textit{GSM8K} \cite{cobbe2021training} — a critical gap remains in evaluating their capacity to understand, analyze, and grade human-generated solutions according to a predefined rubric. This paper addresses this gap by introducing a novel benchmark derived from the Russian Unified State Exam (EGE), a high-stakes graduation examination with a specialized mathematics track.

The EGE mathematics exam comprises 19 tasks, with the second part requiring detailed, handwritten solutions that are manually graded by expert teachers. These experts assess scanned solutions based on clearly defined criteria that evaluate not just the final answer, but the correctness of intermediate steps and the validity of the underlying reasoning. Our benchmark leverages this structure by compiling a dataset of 122 problem solutions from the official EGE expert guide, including scanned handwritten work and the corresponding expert-assigned grades. This allows us to evaluate Vision-Language Models (VLMs) on their ability to emulate expert graders, a task that demands a shift from problem-solving to solution-assessment.

We believe this approach offers a highly innovative perspective on model evaluation. Instead of merely testing a model's problem-solving capabilities, our benchmark probes its capacity for nuanced understanding of human thought processes, error identification, and adherence to structured assessment rubrics. This is crucial for developing AI systems that can genuinely assist in educational assessment. Recent research has highlighted the challenges VLMs face, from error propagation due to inaccurate handwriting recognition to a lack of alignment with specific curricula \cite{kasneci2023chatgpt}. By focusing on a single, curriculum-constrained national exam, our work provides a practical case study in bridging the gap between the general capabilities of AI and the specific demands of real-world educational assessment.

We evaluate seven state-of-the-art VLMs across three distinct evaluation modes, analyzing their performance on multiple metrics. Our findings shed light on the current strengths and weaknesses of these models in handling the complexities of handwritten mathematical solutions and the intricacies of expert grading criteria. This research contributes to the broader fields of educational technology and AI, paving the way for more sophisticated and human-centric AI assessment tools.

\section{Related Work}

Our work is situated at the intersection of automated mathematical assessment, multimodal reasoning, and diagnostic evaluation. The field has evolved significantly from its early reliance on deterministic systems to the current exploration of sophisticated AI models.

\subsection{From Answer Verification to Process Analysis}
The earliest forms of automated mathematical assessment were built on Computer Algebra Systems (CAS), with platforms like STACK \cite{sangwin2014computer} excelling at verifying the symbolic equivalence of a final answer. While effective for summative assessments, these systems are fundamentally "correctness-focused" and cannot evaluate the student's reasoning process. This limitation spurred the adoption of Machine Learning and NLP techniques to analyze the textual content of student explanations, often by comparing a student's response to a single "golden" solution. However, this still falls short of diagnosing errors in novel or unexpected solution paths. This long-standing tension between evaluating outcomes versus processes motivates the current research frontier of process-focused assessment, where our work is firmly situated.

\subsection{The Multimodal Challenge of Handwritten Solutions}
Assessing handwritten mathematics is an inherently multimodal task. A specialized field, Handwritten Mathematical Expression Recognition (HMER), has focused on the modular task of transcribing visual notation into a structured format like LaTeX \cite{deng2017watch}. This is a non-trivial problem due to the two-dimensional structure of mathematical expressions and visual ambiguity between symbols.

In parallel, the rise of end-to-end Vision-Language Models (VLMs) like GPT-4o \cite{openai2024o4} has introduced a more integrated paradigm. However, studies applying general-purpose VLMs to grade handwritten assignments have consistently highlighted the problem of \textbf{error propagation}: inaccurate Optical Character Recognition (OCR) of handwriting leads to faulty input for the reasoning module, resulting in incorrect grades \cite{kasneci2023chatgpt}. This suggests that a VLM's generalist vision encoder may be less robust for the specific domain of mathematical notation than a specialized HMER model, making this a critical area for benchmarking.

\subsection{The Paradigm Shift to Diagnostic Assessment}
The most advanced research in this domain has shifted from simply assigning a score to performing diagnostic assessment—identifying the specific nature of a student's error. This requires a model to parse a multi-step solution, compare it to valid reasoning pathways, and classify deviations into a meaningful taxonomy of error types.

This paradigm shift is embodied by the recent development of specialized benchmarks. The \textbf{Fermat} benchmark \cite{yuan2024fermat} was explicitly designed to evaluate a VLM's ability to perform error detection, localization, and correction on handwritten solutions containing synthetically generated, human-verified errors. Similarly, the \textbf{MathCCS} benchmark \cite{wu2024mathccs} uses real-world student data to focus on systematic error analysis as a foundation for generating pedagogically useful feedback. The creation of these rigorous benchmarks signifies a maturation of the field, moving the central question from "Can the model get the score right?" to "Can the model identify and explain the error?" Our work aligns with this trajectory by requiring models to assess solutions against a multi-point rubric that implicitly requires error diagnosis.

\subsection{Evaluating the Evaluators: Reasoning Benchmarks}
The performance of any assessment system is capped by the reasoning power of its underlying models. While text-based benchmarks like \textit{MATH} \cite{hendrycks2021measuring} and \textit{GSM8K} \cite{cobbe2021training} drove initial progress, they suffer from issues like data contamination and an inability to penalize flawed reasoning that leads to a correct answer.

The need to evaluate reasoning in visual contexts has led to more robust multimodal benchmarks. \textbf{MathVista} \cite{lu2023mathvista}, for example, provides a comprehensive suite of problems requiring the interpretation of charts, diagrams, and figures. The significant performance gap between state-of-the-art models and humans on such benchmarks demonstrates that visually-grounded mathematical reasoning remains a formidable challenge. This provides essential context for our work, as it establishes a realistic upper bound on the expected performance of VLMs on the even more complex task of grading, which requires not only solving a problem but diagnosing errors in another agent's solution.

\section{Benchmark Design and Dataset}

Our benchmark is designed to evaluate Vision-Language Models (VLMs) on their ability to assess handwritten mathematical solutions, a task that requires a deep understanding of both visual information and mathematical reasoning. The core of our benchmark is a unique dataset derived from the Russian Unified State Exam (EGE), specifically focusing on the second part of the mathematics exam, where students provide detailed, handwritten solutions.

\subsection{Dataset Sourcing and Characteristics}

The dataset comprises 122 problem solutions, meticulously sourced from the official EGE expert guide. This guide provides a rich collection of real student solutions, along with expert-assigned grades and detailed justifications for those grades. Each entry in our dataset includes:
\begin{itemize}
    \item \textbf{Scanned Handwritten Solution:} An image of the students complete handwritten solution, often spanning multiple pages, capturing the nuances of human handwriting, diagrams, and mathematical notation.
    \item \textbf{Problem Statement:} The original text of the mathematical problem, providing context for the solution.
    \item \textbf{Expert Grade:} The official score assigned by human experts according to the EGE grading criteria.
    \item \textbf{Reference-Based Expert Evaluation:} Includes the final score assigned by a human expert. The assessment is based on a provided \textit{gold-standard} solution and a granular grading rubric, which are available for each task to ensure a transparent and replicable evaluation process.
\end{itemize}

The solutions cover a range of mathematical topics typically found in EGE, including algebra, geometry, trigonometry, and calculus, ensuring a diverse set of challenges for the evaluated models. The handwritten nature of the solutions introduces significant variability in terms of handwriting styles, penmanship, and layout, requiring robust VLM capabilities for accurate interpretation.

\subsection{Mathematical Domains and Task Types}\label{sec:domains}

Each task corresponds to a standard EGE problem type requiring a written solution with reasoning. Table~\ref{tab:task-breakdown} provides an overview of the tasks, including their domain, a brief description, the number of solution samples in our dataset, and the score range (points) for each task.

\vspace{-0.3em}
\begin{table}[t]
\small
\caption{Benchmark breakdown by task type.}
\label{tab:task-breakdown}
\centering
\begin{tabularx}{\linewidth}{@{}lXcc@{}}
\toprule
\textbf{Task ID} & \textbf{Domain} & \textbf{Count} & \textbf{Score Range}\\
\midrule
13 & Trigonometric equations            & 21 & 0--2\\
14 & Stereometry                        & 18 & 0--3\\
15 & Logarithmic inequalities           & 19 & 0--2\\
16 & Financial mathematics problems     & 17 & 0--2\\
17 & Planimetry                         & 15 & 0--3\\
18 & Parameterised equations            & 16 & 0--4\\
19 & Number theory / combinatorics      & 16 & 0--4\\
\bottomrule
\end{tabularx}
\end{table}

\FloatBarrier            % текст пойдёт после таблицы

\subsection{Grading Criteria and Assessment Focus}

The central point of the EGE assessment process is the clearly defined grading criteria for each task. These criteria specify how points are awarded or deducted based on the correctness of the solution steps, the validity of the reasoning, and the accuracy of the final answer. Our benchmark leverages these criteria as the ground truth for evaluation. The primary focus is not on whether the model can solve the problem itself, but rather on its ability to:
\begin{itemize}
    \item \textbf{Understand the Solution Flow:} Comprehend the logical progression of the students solution, including intermediate steps and derivations.
    \item \textbf{Identify Errors:} Accurately pinpoint mathematical errors, logical flaws, or omissions within the handwritten solution.
    \item \textbf{Apply Grading Rubrics:} Assess the identified errors and correct parts of the solution against the specific EGE grading criteria to assign an appropriate score.
\end{itemize}

This emphasis on assessment rather than problem-solving distinguishes our benchmark from many existing math-focused datasets and provides a more realistic evaluation of AI potential in educational grading scenarios.

\section{Experimental Setup}

To evaluate the performance of Vision-Language Models on our EGE-Math Solutions Assessment Benchmark, we conducted experiments with seven different state-of-the-art models. The evaluation was structured around three distinct procedures, or "modes", designed to assess the models' capabilities under different levels of contextual information. This required a meticulous data curation process where a specific version of the dataset was prepared for each mode.

\subsection{Evaluated Models}\label{sec:evaluated-models}

We selected a diverse set of VLMs to cover a range of architectures and capabilities:
\begin{itemize}
    \item \textbf{Arcee AI Spotlight:} A model from Arcee AI, accessed via OpenRouter. \cite{arcee2025blog}
    \item \textbf{Google Gemini 2.0 Flash:} Google's VLM, known for its multimodal capabilities \cite{gemini2023}.
    \item \textbf{Google Gemini 2.0 Flash Lite:} A lighter version of Google Gemini 2.0 Flash.
    \item \textbf{Google Gemini 2.5 Flash Preview:} A preview version of Google's next-generation VLM.
    \item \textbf{Google Gemini 2.5 Flash Preview:thinking:} A variant of Google's Gemini 2.5 Flash Preview with enhanced reasoning abilities.
    \item \textbf{OpenAI o4-mini:} A model from OpenAI, a smaller, more efficient version of their flagship models. \cite{openai2025introducing-o4mini}
    \item \textbf{Qwen 2.5 VL 32B:} A large Vision-Language Model from Alibaba Cloud, accessed via OpenRouter \cite{bai2025qwen25vltechnicalreport}.
\end{itemize}

Each model was prompted to analyze the handwritten solution image and provide an assessment based on the EGE grading criteria. The output format was standardized to facilitate automated comparison with expert grades.

\subsection{Evaluation Procedure and Data Curation}\label{sec:procedure-curation}

To thoroughly test the model's understanding and reasoning, we designed and prepared data for three evaluation modes. This approach allows for a granular analysis of how additional context influences the models' assessment performance.

\begin{itemize}
    \item \textbf{Mode 1: Without Answer.} In this mode, the model receives only the handwritten solution image and the problem statement. To facilitate this, we prepared a \textbf{baseline dataset} where each entry consisted of a pre-processed image and the problem text. The image pre-processing involved standardizing dimensions and resolution to ensure consistent input quality across all experiments. This mode assesses the model's ability to assign a grade based solely on the provided content and its internal understanding of the EGE grading rubric.

    \item \textbf{Mode 2: With Answer.} For this mode, the model receives the handwritten solution, the problem statement, and the correct final numerical answer. To enable this, the baseline dataset was \textbf{augmented} by appending the correct final answer for each of the 122 problems, sourced from official EGE materials. This mode assesses whether the model can leverage the correct outcome to better identify errors or confirm the correctness of the student’s solution steps.

    \item \textbf{Mode 3: With True Solution.} This is the most informative mode, where the model is given the handwritten solution, the problem statement, and a complete, correct reference solution. The dataset for this mode was \textbf{further enriched} with a transcribed, step-by-step "gold standard" solution from the EGE expert guide. This allows us to evaluate the model's ability to compare the student’s approach with a known correct method and identify deviations or errors more precisely.
\end{itemize}

\subsection{Prompt Templates and Score Extraction Methodology}\label{sec:prompts}

For each evaluation, the models were provided with specific prompt templates tailored to the task and evaluation mode. These templates included the problem description, the student's handwritten solution (as an image), and the relevant grading criteria. For the \textbf{With Answer} and \textbf{With True Solution} modes, the correct answer or reference solution was also incorporated into the prompt. The models were instructed to output their assessment in the structured format, including the analysis of the solution, the final score, and the justification for that score. This structured output facilitated automated extraction of the assigned scores for quantitative analysis. The full prompt templates used for all evaluation modes are available in the project's public repository.

\section{Results}

Our evaluation of seven Vision-Language Models across three distinct evaluation modes provides insights into their capabilities in assessing handwritten mathematical solutions.

\subsection{Metrics}
We report three complementary metrics:

\paragraph{Accuracy (Exact Match:)} Percentage of cases where the predicted score exactly matches the expected score:
\[
\text{Accuracy} = \frac{\text{Exact Matches}}{\text{Total Evaluations}}\times100\%.
\]

\paragraph{Quality Score:} Normalized closeness between predicted and expected scores:
\[
\text{Quality Score} = 100\%\times\Bigl(1-\frac{|S_{\mathrm{pred}}-S_{\mathrm{true}}|}{S_{\mathrm{max}}}\Bigr),
\]
where $S_{\mathrm{max}}\in\{2,3,4\}$ is the task-specific maximum.

\paragraph{Average Score Distance:}
\[
\text{Avg.\ Distance}=\frac{1}{n}\sum_{i=1}^{n}\bigl|S_{\mathrm{pred},i}-S_{\mathrm{true},i}\bigr|.
\]

\subsection{Performance Analysis}

As can be seen from Table \ref{tab:overall_results}, OpenAI o4-mini consistently demonstrates the highest performance across all evaluation modes, achieving the best Accuracy (56.56\% with Answer) and Quality Score (78.17\% with Answer), and the lowest Average Score Distance (0.60 with Answer). This suggests that OpenAI\textquotesingle s model possesses superior capabilities in understanding handwritten solutions and applying grading criteria compared to other evaluated models. 

Among other models, Google Gemini 2.0 Flash also shows strong performance, particularly in the \textbf{With Answer} and \textbf{With True Solution} modes, indicating its ability to effectively leverage additional context. Models like Arcee AI Spotlight and Qwen 2.5 VL 32B exhibit lower accuracy and higher score distances, suggesting that while they can process the visual input, their mathematical reasoning and grading alignment are less precise. The \textit{thinking} variant of Google Gemini 2.5 Flash Preview, despite its higher cost and longer average time, does not consistently outperform its non-\textit{thinking} counterpart, raising questions about the efficacy of its enhanced reasoning capabilities for this specific task. To offer a qualitative perspective on these quantitative results, we present a detailed case study of a single solution's assessment in Appendix, which illustrates the models' divergent failure modes.

\begin{figure}[t]
\centering

\includegraphics[width=\linewidth]{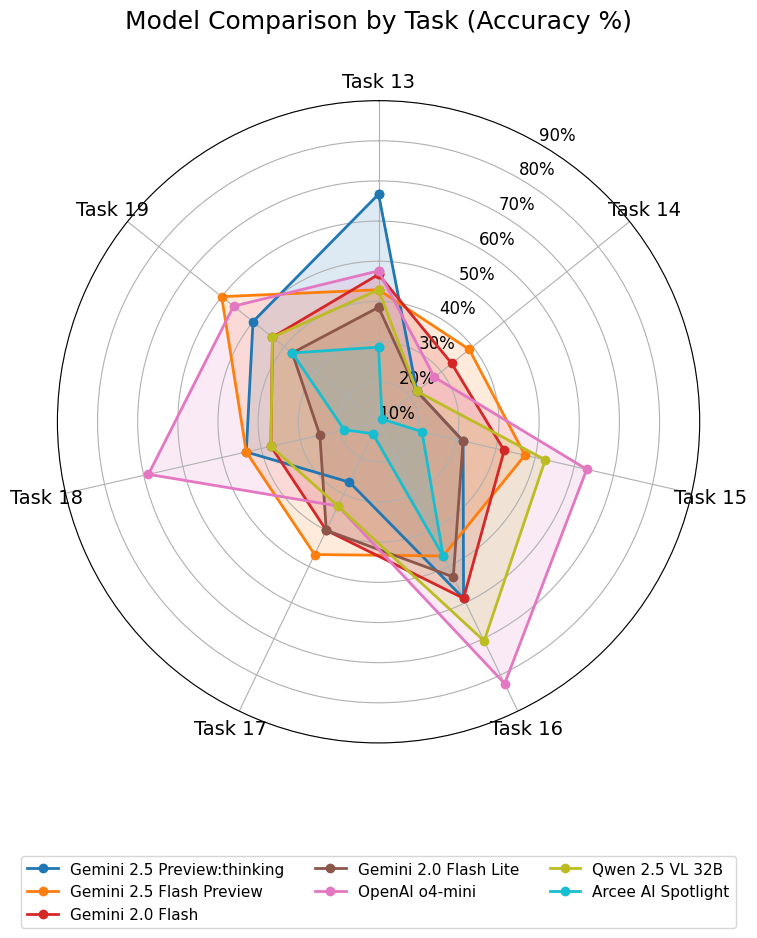} 
\caption{Radar chart showing model Accuracy (\%) in the With True Solution mode across all seven task types. The outer edge represents a perfect score. This visualization highlights the models' strengths and weaknesses on different mathematical domains.}
\label{fig:radar_chart}
\end{figure}

A detailed breakdown of performance by task type, illustrated in Figure~\ref{fig:radar_chart}, reveals significant variations. It is evident that algebraic tasks (13 and 15) are handled more effectively by most models. In contrast, both geometry categories (14 — stereometry, 17 — planimetry) consistently yield poorer agreement with human graders. We hypothesise that current VLMs still struggle to map free-hand diagrams onto the rigorous spatial reasoning chains required by the EGE rubric. The full per-task scores for all models can be found in Appendix.

\begin{table*}[t]
\centering
\caption{Overall performance of all models across three evaluation modes. The best result for each combination of mode and metric is shown in bold, and the second best result is underlined.}
\label{tab:overall_results}
\scriptsize
\setlength\tabcolsep{4pt}
\begin{tabular}{l l l c c c c c}
\toprule
\textbf{Model} & \textbf{Provider} & \textbf{Mode} & \textbf{Acc. (\%)} & \textbf{Qual. (\%)} & \textbf{Avg. Dist.} & \textbf{Cost (\$)} & \textbf{Time (s)} \\
\midrule
\multirow{3}{*}{Arcee AI Spotlight} & \multirow{3}{*}{Arcee AI (via OpenRouter)} & Without Answer & 27.87 & 64.48 & 1.04 & \textbf{<0.01} & 8.80 \\
& & With Answer & 26.23 & 63.18 & 1.09 & \textbf{<0.01} & 6.99 \\
& & With True Solution & 25.41 & 59.22 & 1.16 & \textbf{<0.01} & 6.98 \\
\midrule
\multirow{3}{*}{Google Gemini 2.0 Flash} & \multirow{3}{*}{Google} & Without Answer & 36.89 & \underline{71.04} & 0.84 & 0.14 & \underline{4.56} \\
& & With Answer & \underline{47.54} & \underline{74.04} & \underline{0.75} & 0.14 & \underline{4.82} \\
& & With True Solution & \underline{46.72} & \underline{75.82} & \underline{0.71} & 0.21 & \underline{3.13} \\
\midrule
\multirow{3}{*}{Google Gemini 2.0 Flash Lite} & \multirow{3}{*}{Google} & Without Answer & 31.97 & 64.96 & 1.00 & \underline{0.04} & \textbf{3.08} \\
& & With Answer & 35.25 & 67.83 & 0.90 & \underline{0.04} & \textbf{3.13} \\
& & With True Solution & 38.52 & 70.22 & 0.84 & \underline{0.04} & \textbf{3.09} \\
\midrule
\multirow{3}{*}{Google Gemini 2.5 Flash Preview} & \multirow{3}{*}{Google} & Without Answer & \underline{44.26} & \underline{71.04} & \underline{0.81} & 0.32 & 16.08 \\
& & With Answer & 40.98 & 70.49 & 0.82 & 0.30 & 14.92 \\
& & With True Solution & 45.90 & 71.35 & 0.79 & 0.34 & 11.67 \\
\midrule
\multirow{3}{*}{Google Gemini 2.5 Flash Preview:thinking} & \multirow{3}{*}{Google} & Without Answer & 40.16 & 64.30 & 1.05 & 0.60 & 39.48 \\
& & With Answer & 42.62 & 66.44 & 0.99 & 0.62 & 39.98 \\
& & With True Solution & 43.44 & 65.92 & 0.99 & 0.78 & 47.59 \\
\midrule
\multirow{3}{*}{OpenAI o4-mini} & \multirow{3}{*}{OpenAI} & Without Answer & \textbf{55.74} & \textbf{75.55} & \textbf{0.66} & 2.18 & 39.62 \\
& & With Answer & \textbf{56.56} & \textbf{78.17} & \textbf{0.60} & 2.02 & 32.94 \\
& & With True Solution & \textbf{54.10} & \textbf{76.16} & \textbf{0.66} & 2.28 & 58.47 \\
\midrule
\multirow{3}{*}{Qwen 2.5 VL 32B} & \multirow{3}{*}{Alibaba Cloud (via OpenRouter)} & Without Answer & 31.15 & 62.09 & 1.09 & 0.46 & 22.97 \\
& & With Answer & 30.33 & 61.95 & 1.08 & 0.46 & 23.27 \\
& & With True Solution & 43.44 & 70.49 & 0.81 & 0.63 & 27.55 \\
\bottomrule
\end{tabular}
\end{table*}

\subsection{Impact of Evaluation Modes}

One of the most interesting findings is the varied impact of the evaluation modes on model performance. For some models, providing additional context (correct answer or true solution) significantly improved their performance. For instance, Google Gemini 2.0 Flash showed a notable increase in Accuracy when provided with the correct answer (from 36.89\% to 47.54\%). This suggests that these models can effectively leverage external information to refine their assessment, indicating a capacity for conditional reasoning. However, this improvement was not universal; Arcee AI Spotlight, for example, saw a slight decrease in performance with additional context, which might indicate issues with how it integrates or prioritizes external information versus its internal analysis of the handwritten solution.

The \textbf{With True Solution} mode, while providing the most comprehensive context, did not consistently lead to the best performance across all models. This could be attributed to several factors: the models might struggle with effectively comparing a student\textquotesingle s potentially divergent solution path with a provided reference solution, or lack the complexity sufficient to fully leverage the detailed information in a reference solution when the student\textquotesingle s approach deviates significantly. This highlights a crucial area for future research: developing VLMs that can perform robust comparative analysis between a student\textquotesingle s solution and a reference, even when the two solution paths differ.

% \section{Discussion}

% Our evaluation provides a unique perspective on the capabilities of Vision-Language Models in a challenging, real-world assessment scenario. The EGE-Math Solutions Assessment Benchmark highlights that while current VLMs have made significant strides in multimodal understanding, there remains a substantial gap in their ability to perform nuanced, criterion-referenced grading of handwritten mathematical solutions at a human expert level. The highest accuracy achieved (56.56\%) indicates that there is ample room for improvement, particularly in complex mathematical reasoning and the precise application of grading rubrics.

\section{Discussion and Limitations}

Our evaluation provides a unique perspective on VLM capabilities in a real-world assessment scenario. The results highlight a substantial gap between current model performance and human expert-level grading, with the highest accuracy at 56.56\%. This indicates ample room for improvement in nuanced mathematical reasoning and the precise application of grading rubrics.

Several factors contribute to these performance limitations and pave the way for future research:

\begin{itemize}
    \item \textbf{Visual Interpretation and Error Propagation:} The diversity in student handwriting, penmanship, and layout poses a significant challenge for accurate visual interpretation. This often leads to \textit{error propagation}, where inaccuracies in the initial visual recognition are passed downstream to the reasoning module, causing incorrect assessments. Future work could explore hybrid approaches, combining general VLM perception with specialized Handwritten Mathematical Expression Recognition (HMER) models to mitigate this issue.

    \item \textbf{Deep Reasoning and Rubric Alignment:} Assessing complex solutions requires deep symbolic and logical reasoning, especially for non-standard solution paths or subtle errors. Models often struggle to translate qualitative grading criteria into quantitative scores, sometimes misinterpreting the severity of an error or failing to identify all relevant mistakes.

    \item \textbf{Dataset and Fine-Tuning:} The current benchmark utilizes 122 solutions. A larger, more diverse dataset would enable more comprehensive evaluation. Furthermore, our study primarily relies on zero-shot prompting; fine-tuning VLMs on this specific assessment task could significantly improve their performance and alignment with the specific curriculum and rubrics.

    \item \textbf{Contextual Reasoning:} While some models effectively use additional context (like a correct answer), others struggle to integrate this information. This highlights a need for more robust mechanisms for conditional reasoning and information fusion in VLMs.

    \item \textbf{Explainability and Future Directions:} The transparency and interpretability of the models' reasoning processes remain a challenge. Developing more explainable AI is crucial for building trust and utility in educational assessment tools. Future work could also explore interactive assessment scenarios or adapt the benchmark to other global curricula to test for generalization.
\end{itemize}

% \subsection{Challenges and Limitations}
% Several factors contribute to the observed performance limitations. The diverse handwriting styles, penmanship, and layouts in students' solutions pose a significant challenge for accurate visual interpretation. Furthermore, assessing complex mathematical solutions requires not just pattern recognition, but deep symbolic and logical reasoning, which current VLMs may still struggle with, especially when faced with non-standard solution paths or subtle errors. Translating qualitative grading criteria into quantitative scores also demands a sophisticated understanding of mathematical concepts and the ability to weigh different types of errors according to the rubric, where models may misinterpret the severity of an error or fail to identify all relevant errors. Finally, while some models benefit from additional context, others struggle to effectively integrate this information, suggesting a need for more robust mechanisms for conditional reasoning and information fusion in VLMs.

\subsection{Cost and Efficiency}

Beyond performance, our study also sheds light on the practical considerations of deploying such models for automated assessment. The significant variation in total cost and average evaluation time across models (e.g., OpenAI o4-mini being considerably more expensive and slower than Google Gemini 2.0 Flash Lite) highlights a trade-off between performance and operational efficiency. For large-scale deployment in educational settings, cost-effectiveness and speed are critical factors that need to be balanced against grading accuracy. Larger frontier models (e.g.\ OpenAI \texttt{o3} or Google \texttt{gemini-2.5-pro}) were not included in this benchmark due to computational and budgetary constraints; their evaluation remains future work.

% \section{Limitations}
% Our study, while providing valuable insights, has certain limitations that pave the way for future research. The current benchmark utilizes 122 solutions; a larger dataset with a wider variety of problems, solution types, and error patterns would enable a more comprehensive evaluation and potentially improve model training for fine-tuning. Our evaluation primarily uses zero-shot or few-shot prompting, and fine-tuning VLMs on this specific assessment task could significantly improve their performance. Additionally, exploring hybrid approaches that combine direct visual understanding with text extracted via Optical Character Recognition (OCR) could be a fruitful direction. The transparency and interpretability of the models' reasoning processes also remain a challenge, highlighting the need for more explainable AI systems in educational assessment. Future work could also explore interactive assessment scenarios where models provide iterative feedback or engage in a dialogue with the student or educator. Finally, while the EGE provides a rich context, this benchmark could be adapted to other curricula and grading systems globally.

\section{License}
The source code and dataset for this research are available under the MIT License. This permissive license allows for reuse, modification, and distribution, both in academic and commercial settings, provided that the original copyright and license notice are included.

\section{Conclusion}

This paper introduces a novel and challenging benchmark for evaluating Vision-Language Models on the task of assessing handwritten mathematical solutions from the Russian Unified State Exam. Our findings demonstrate that while state-of-the-art VLMs can perform this complex task, there is significant room for improvement in terms of accuracy, nuanced reasoning, and cost-efficiency. The benchmark highlights the critical need for VLMs that can robustly interpret diverse handwriting, apply intricate grading rubrics, and effectively integrate external contextual information. This research contributes to the growing field of AI in education, providing a valuable tool for developing and evaluating next-generation AI-assisted assessment systems that can truly support students and educators in the complex domain of mathematical learning and evaluation.

\bibliography{references}

\begin{thebibliography}{13}
\providecommand{\natexlab}[1]{#1}

\bibitem[{{Arcee.ai}(2025)}]{arcee2025blog}
{Arcee.ai}. 2025.
\newblock Arcee Blog.
\newblock \url{https://www.arcee.ai/blog}.
\newblock Accessed: 2025-07-06.

\bibitem[{Bai et~al.(2025)Bai, Chen, Liu, Wang, Ge, Song, Dang, Wang, Wang, Tang, Zhong, Zhu, Yang, Li, Wan, Wang, Ding, Fu, Xu, Ye, Zhang, Xie, Cheng, Zhang, Yang, Xu, and Lin}]{bai2025qwen25vltechnicalreport}
Bai, S.; Chen, K.; Liu, X.; Wang, J.; Ge, W.; Song, S.; Dang, K.; Wang, P.; Wang, S.; Tang, J.; Zhong, H.; Zhu, Y.; Yang, M.; Li, Z.; Wan, J.; Wang, P.; Ding, W.; Fu, Z.; Xu, Y.; Ye, J.; Zhang, X.; Xie, T.; Cheng, Z.; Zhang, H.; Yang, Z.; Xu, H.; and Lin, J. 2025.
\newblock Qwen2.5-VL Technical Report.
\newblock arXiv:2502.13923.

\bibitem[{Cobbe et~al.(2021)Cobbe, Kosaraju, Bavarian, Chen, Jun, Kaiser, Plappert, Tworek, Hilton, Nakano et~al.}]{cobbe2021training}
Cobbe, K.; Kosaraju, V.; Bavarian, M.; Chen, M.; Jun, H.; Kaiser, L.; Plappert, M.; Tworek, J.; Hilton, J.; Nakano, R.; et~al. 2021.
\newblock Training Verifiers to Solve Math Word Problems.
\newblock \emph{arXiv preprint arXiv:2110.14168}.

\bibitem[{Deng et~al.(2017)Deng, Kan, Yin, and Zhang}]{deng2017watch}
Deng, Y.; Kan, A.; Yin, F.; and Zhang, Z. 2017.
\newblock Watch, attend and parse: An end-to-end neural network based approach to handwritten mathematical expression recognition.
\newblock \emph{Pattern Recognition}, 71: 196--206.

\bibitem[{Hendrycks et~al.(2021)Hendrycks, Burns, Kadavath, Arora, Basart, Tang, Song, and Steinhardt}]{hendrycks2021measuring}
Hendrycks, D.; Burns, C.; Kadavath, S.; Arora, A.; Basart, S.; Tang, E.; Song, D.; and Steinhardt, J. 2021.
\newblock Measuring Mathematical Problem Solving with the MATH Dataset.
\newblock \emph{arXiv preprint arXiv:2103.03874}.

\bibitem[{Kasneci et~al.(2023)Kasneci, Se{\ss}ler, K{\"u}chemann, Bannert, Dementieva, Fischer, Gasser, Groh, G{\"u}nnemann, H{\"u}llermeier et~al.}]{kasneci2023chatgpt}
Kasneci, E.; Se{\ss}ler, K.; K{\"u}chemann, S.; Bannert, M.; Dementieva, D.; Fischer, F.; Gasser, U.; Groh, G.; G{\"u}nnemann, S.; H{\"u}llermeier, E.; et~al. 2023.
\newblock ChatGPT for good? On opportunities and challenges of large language models for education.
\newblock \emph{Learning and individual differences}, 103: 102274.

\bibitem[{Lu et~al.(2023)Lu, Bansal, Xia, Liu, Li, Hajishirzi, Cheng, Chang, Galley, and Gao}]{lu2023mathvista}
Lu, P.; Bansal, H.; Xia, T.; Liu, J.; Li, C.; Hajishirzi, H.; Cheng, H.; Chang, K.-W.; Galley, M.; and Gao, J. 2023.
\newblock MathVista: Evaluating Mathematical Reasoning of Foundation Models in Visual Contexts.
\newblock \emph{arXiv preprint arXiv:2310.02255}.

\bibitem[{OpenAI(2024)}]{openai2024o4}
OpenAI. 2024.
\newblock GPT-4o.

\bibitem[{{OpenAI}(2025)}]{openai2025introducing-o4mini}
{OpenAI}. 2025.
\newblock Introducing O3 and O4‑mini.
\newblock \url{https://openai.com/index/introducing-o3-and-o4-mini/}.
\newblock Accessed: 2025‑07‑06.

\bibitem[{Sangwin(2014)}]{sangwin2014computer}
Sangwin, C.~J. 2014.
\newblock Computer-aided assessment of mathematics using STACK.
\newblock \emph{ZDM}, 46(2): 307--320.

\bibitem[{Team et~al.(2023)}]{gemini2023}
Team, G.; et~al. 2023.
\newblock Gemini: A Family of Highly Capable Multimodal Models.
\newblock \emph{arXiv preprint arXiv:2312.11805}.

\bibitem[{Wu et~al.(2024)Wu, Li, Li, and Zhou}]{wu2024mathccs}
Wu, Y.; Li, Y.; Li, Y.; and Zhou, W. 2024.
\newblock MathCCS: A New Benchmark for Mathematical Classification and Constructive Suggestions.
\newblock \emph{arXiv preprint arXiv:2405.17642}.

\bibitem[{Yuan et~al.(2024)Yuan, Zhang, Liu, Wang, Zhang, Liu, and Chua}]{yuan2024fermat}
Yuan, Z.; Zhang, Y.; Liu, J.; Wang, Y.; Zhang, J.; Liu, H.; and Chua, T.-S. 2024.
\newblock Fermat: A Benchmark for Evaluating VLM's Ability in Factual Error Correction of Handwritten Math Solutions.
\newblock \emph{arXiv preprint arXiv:2405.10100}.

\end{thebibliography}
% \bibliographystyle{aaai2026}
% \bibliography{aaai2026}
% \setcounter{secnumdepth}{2}
\clearpage

\onecolumn
\appendix
\section{Appendix: Per-Task Performance Data}

This appendix provides the detailed per-task scores for a selection of the evaluated models.

\begin{table}[H]
\scriptsize
\centering
\caption{Per-task scores — \texttt{openai\_o4-mini}.}
\label{tab:openai-mini}
\begin{tabular*}{\linewidth}{@{\extracolsep{\fill}}lccccc@{}}
\toprule
Task & Examples & Accuracy & Average Score & Expected Score & Cost\\
\midrule
13 & 21 & 47.6\% & 1.48 & 0.95 & \$0.4259 \\
14 & 18 & 27.8\% & 1.72 & 1.28 & \$0.3465 \\
15 & 19 & \textbf{63.2\%} & 1.68 & 1.11 & \$0.3115 \\
16 & 17 & \textbf{82.4\%} & 1.24 & 1.29 & \$0.2957 \\
17 & 15 & 33.3\% & 1.20 & 1.20 & \$0.2560 \\
18 & 16 & \textbf{68.8\%} & 2.12 & 2.38 & \$0.3543 \\
19 & 16 & \textbf{56.2\%} & 1.75 & 2.06 & \$0.2879 \\
\bottomrule
\end{tabular*}
\end{table}

\begin{table}[H]
\scriptsize
\centering
\caption{Per-task scores — \texttt{qwen-2.5-vl-32b}.}
\label{tab:qwen-vl}
\begin{tabular*}{\linewidth}{@{\extracolsep{\fill}}lccccc@{}}
\toprule
Task & Examples & Accuracy & Average Score & Expected Score & Cost\\
\midrule
13 & 21 & 42.9\% & 1.62 & 0.95 & \$0.1095 \\
14 & 18 & 22.2\% & 2.17 & 1.28 & \$0.0999 \\
15 & 19 & \textbf{52.6\%} & 1.58 & 1.11 & \$0.0875 \\
16 & 17 & \textbf{70.6\%} & 1.41 & 1.29 & \$0.0783 \\
17 & 15 & 33.3\% & 1.73 & 1.20 & \$0.0753 \\
18 & 16 & 37.5\% & 2.75 & 2.38 & \$0.0970 \\
19 & 16 & 43.8\% & 2.06 & 2.06 & \$0.0868 \\
\bottomrule
\end{tabular*}
\end{table}

\begin{table}[H]
\scriptsize
\centering
\caption{Per-task scores — \texttt{arcee-ai\_spotlight}.}
\label{tab:arcee-spotlight}
\begin{tabular*}{\linewidth}{@{\extracolsep{\fill}}lccccc@{}}
\toprule
Task & Examples & Accuracy & Average Score & Expected Score & Cost\\
\midrule
13 & 21 & 28.6\% & 0.95 & 0.95 & $< \$0.01$ \\
14 & 18 & 11.1\% & 2.72 & 1.28 & $< \$0.01$ \\
15 & 19 & 21.1\% & 0.74 & 1.11 & $< \$0.01$ \\
16 & 17 & 47.1\% & 1.47 & 1.29 & $< \$0.01$ \\
17 & 15 & 13.3\% & 2.67 & 1.20 & $< \$0.01$ \\
18 & 16 & 18.8\% & 2.12 & 2.38 & $< \$0.01$ \\
19 & 16 & 37.5\% & 2.62 & 2.06 & $< \$0.01$ \\
\bottomrule
\end{tabular*}
\end{table}

\begin{table}[H]
\scriptsize
\centering
\caption{Per-task scores — \texttt{gemini-2.5-flash-preview}.}
\label{tab:gemini-preview}
\begin{tabular*}{\linewidth}{@{\extracolsep{\fill}}lccccc@{}}
\toprule
Task & Examples & Accuracy & Average Score & Expected Score & Cost\\
\midrule
13 & 21 & 42.9\% & 1.48 & 0.95 & \$0.0493 \\
14 & 18 & 38.9\% & 1.00 & 1.28 & \$0.0616 \\
15 & 19 & 47.4\% & 1.79 & 1.11 & \$0.0401 \\
16 & 17 & 47.1\% & 1.41 & 1.29 & \$0.0419 \\
17 & 15 & 46.7\% & 0.73 & 1.20 & \$0.0387 \\
18 & 16 & 43.8\% & 1.81 & 2.38 & \$0.0713 \\
19 & 15 & \textbf{60.0\%} & 1.40 & 2.13 & \$0.0414 \\
\bottomrule
\end{tabular*}
\end{table}

\begin{table}[H]
\scriptsize
\centering
\caption{Per-task scores — \texttt{gemini-2.5-flash-preview\_thinking}.}
\label{tab:gemini-thinking}
\begin{tabular*}{\linewidth}{@{\extracolsep{\fill}}lccccc@{}}
\toprule
Task & Examples & Accuracy & Average Score & Expected Score & Cost\\
\midrule
13 & 21 & \textbf{66.7\%} & 1.57 & 0.95 & \$0.1096\\
14 & 18 & 22.2\% & 1.33 & 1.28 & \$0.1043\\
15 & 19 & 31.6\% & 1.21 & 1.11 & \$0.0998\\
16 & 17 & \textbf{58.8\%} & 1.53 & 1.29 & \$0.0964\\
17 & 15 & 26.7\% & 0.73 & 1.20 & \$0.0908\\
18 & 16 & 43.8\% & 2.50 & 2.38 & \$0.1037\\
19 & 16 & 50.0\% & 2.44 & 2.06 & \$0.1787\\
\bottomrule
\end{tabular*}
\end{table}

\begin{table}[H]
\scriptsize
\centering
\caption{Per-task scores — \texttt{gemini-2.0-flash-001}.}
\label{tab:gemini-flash}
\begin{tabular*}{\linewidth}{@{\extracolsep{\fill}}lccccc@{}}
\toprule
Task & Examples & Accuracy & Average Score & Expected Score & Cost\\
\midrule
13 & 21 & \textbf{61.9\%} & 1.48 & 0.95 & \$0.0295\\
14 & 18 & 33.3\% & 1.61 & 1.28 & \$0.0284\\
15 & 19 & 42.1\% & 1.42 & 1.11 & \$0.0276\\
16 & 17 & \textbf{58.8\%} & 1.47 & 1.29 & \$0.0270\\
17 & 15 & 40.0\% & 0.93 & 1.20 & \$0.0254\\
18 & 16 & 37.5\% & 2.50 & 2.38 & \$0.0286\\
19 & 16 & 43.8\% & 2.31 & 2.06 & \$0.0392\\
\bottomrule
\end{tabular*}
\end{table}

\begin{table}[H]
\scriptsize
\centering
\caption{Per-task scores — \texttt{gemini-2.0-flash-lite-001}.}
\label{tab:gemini-lite}
\begin{tabular*}{\linewidth}{@{\extracolsep{\fill}}lccccc@{}}
\toprule
Task & Examples & Accuracy & Average Score & Expected Score & Cost\\
\midrule
13 & 21 & \textbf{57.1\%} & 1.38 & 0.95 & \$0.0059\\
14 & 18 & 22.2\% & 1.50 & 1.28 & \$0.0056\\
15 & 19 & 31.6\% & 1.26 & 1.11 & \$0.0052\\
16 & 17 & \textbf{52.9\%} & 1.29 & 1.29 & \$0.0051\\
17 & 15 & 40.0\% & 0.87 & 1.20 & \$0.0048\\
18 & 16 & 25.0\% & 2.62 & 2.38 & \$0.0054\\
19 & 16 & 37.5\% & 2.06 & 2.06 & \$0.0049\\
\bottomrule
\end{tabular*}
\end{table}

\appendix
\section{Appendix: Representative Example Analysis}

This appendix provides a detailed analysis of a representative example from our benchmark to illustrate the evaluation process and the typical performance patterns of the models.

\subsection{Analysis of Solution 18.3.3}

\subsubsection{Problem Statement}
Find all values of parameter $a$ for which the equation has exactly three distinct roots:
\[ \sqrt{3x^2 + 2ax + 1} = x^2 + ax + 1 \]

\subsubsection{Official Grading Criteria (Task 18)}
\begin{itemize}[nosep]
    \item \textbf{4 points:} A well-reasoned, correct solution is provided.
    \item \textbf{3 points:} A set of parameter values is obtained that differs from the correct set only by the inclusion/exclusion of boundary points.
    \item \textbf{2 points:} A correct interval of parameter values is obtained (possibly with incorrect boundary points), OR an incorrect answer is obtained due to a computational error, but all logical steps are correct.
    \item \textbf{1 point:} The roots of the equation are found, and the problem is correctly reduced to investigating these roots under the given condition(s).
    \item \textbf{0 points:} The solution does not meet any of the criteria above.
\end{itemize}

\subsubsection{Visual Materials}
\begin{figure}[H]
    \centering
    \includegraphics[width=0.8\linewidth]{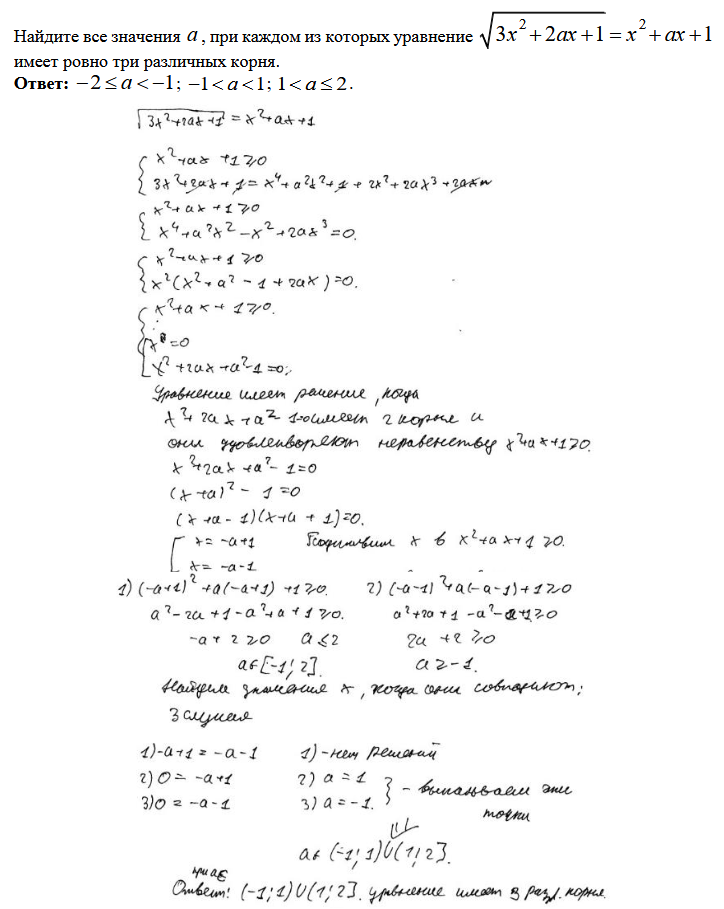}
    \caption{Student's handwritten solution for problem 18.3.3. The expert-assigned score is 2.}
    \label{fig:student_solution}
\end{figure}

\begin{figure}[H]
    \centering
    \includegraphics[width=0.8\linewidth]{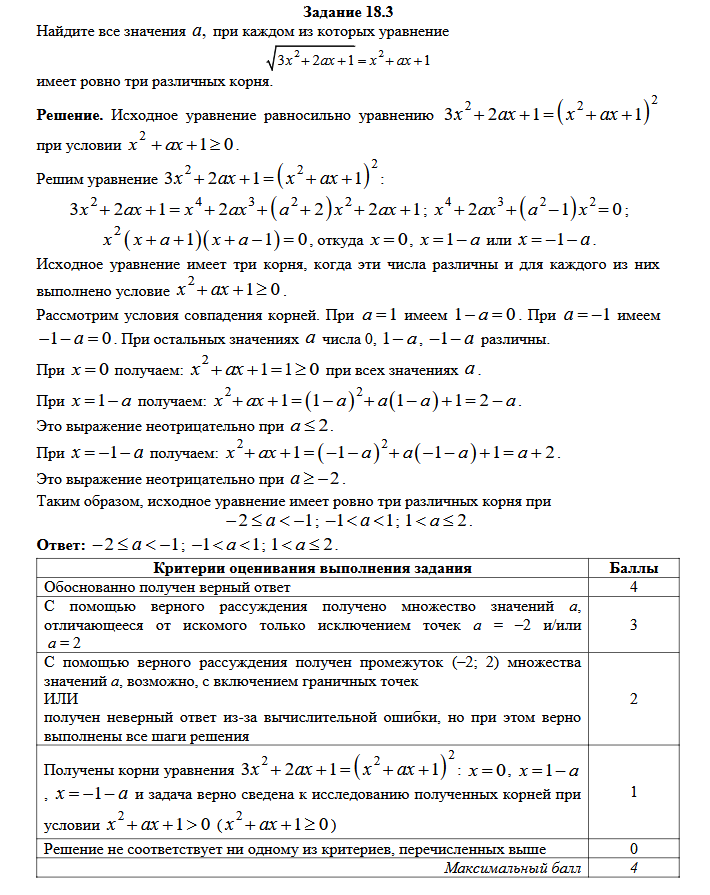}
    \caption{Official correct solution for problem 18.3.}
    \label{fig:true_solution}
\end{figure}

Figure~\ref{fig:student_solution} shows the student's handwritten solution, and Figure~\ref{fig:true_solution} shows the official correct solution provided in the EGE expert guide.

\newpage
\subsection{Model Assessment Results}
The following table summarizes the scores assigned by different models in the \textbf{With True Solution} mode. The expected score was 2.

\begin{table}[H]
\centering
\caption{Model scores for solution 18.3.3.}
\begin{tabular}{lccc}
\toprule
\textbf{Model} & \textbf{Assigned Score} & \textbf{Expected} & \textbf{Result} \\
\midrule
OpenAI o4-mini & \textcolor{green!60!black}{\textbf{2}} & 2 & Correct \\
Qwen 2.5 VL 32B & \textcolor{red}{\textbf{4}} & 2 & Overestimated \\
Google Gemini 2.0 Flash & \textcolor{green!60!black}{\textbf{2}} & 2 & Correct \\
Google Gemini 2.0 Flash Lite & \textcolor{green!60!black}{\textbf{2}} & 2 & Correct \\
Google Gemini 2.5 Flash Preview & \textcolor{green!60!black}{\textbf{2}} & 2 & Correct \\
Google Gemini 2.5 Flash Preview Thinking & \textcolor{green!60!black}{\textbf{2}} & 2 & Correct \\
Arcee-AI Spotlight & \textcolor{red}{\textbf{0}} & 2 & Underestimated \\
\bottomrule
\end{tabular}
\end{table}

\subsubsection{Key Observations}
\begin{itemize}[nosep]
    \item \textbf{High Accuracy of Most Models:} The majority of the models (5 out of 7) successfully handled the task, assigning the correct score of 2. This group included OpenAI o4-mini and all the tested models from the Google Gemini family.
    \item \textbf{Divergent Errors:} Two models evaluated the solution incorrectly, and their errors were opposites. Qwen 2.5 VL 32B significantly overestimated the score (\textbf{4}), while Arcee-AI Spotlight failed to produce a final answer: it became stuck in a loop of writing out equations, which resulted in a score of (\textbf{0}).
    \item \textbf{Distinct Failure Modes:} The errors highlight very different failure modes. One model overestimated the score, while the other failed to complete the task entirely. This points to unique flaws in the logic of each model rather than a shared, systematic bias.
\end{itemize}

\subsection{Full Model Responses and Prompt (Translated to English)}

\subsubsection{Prompt Used for Evaluation (With True Solution)}

\vspace{1ex}

\begin{mdframed}[backgroundcolor=gray!5] 
\small\ttfamily

\subsection*{Analyze the solution of task 18}
(an equation, inequality, or system thereof with a parameter) and evaluate it according to the criteria.

\subsubsection*{Task}
\{task description\}

\subsubsection*{Assessment criteria for task 18}
\begin{itemize}
    \item \textbf{4 points:} A correct answer is obtained with justification.
    \item \textbf{3 points:} A set of parameter values is obtained through correct reasoning, differing from the required set only by the exclusion of boundary points or the inclusion of points not belonging to the answer.
    \item \textbf{2 points:} An interval of the set of parameter values is obtained through correct reasoning, possibly including boundary points, OR an incorrect answer is obtained due to a computational error, but all steps of the solution are correctly performed.
    \item \textbf{1 point:} The roots of the equation are found, and the problem is correctly reduced to the investigation of these roots under the given condition(s).
    \item \textbf{0 points:} The solution does not meet any of the criteria listed above.
\end{itemize}

\subsubsection*{IMPORTANT: Assessment Principles}
\begin{itemize}
    \item Evaluate the solution \textbf{STRICTLY} according to the criteria.
    \item Pay attention to mathematical correctness, not the presentation.
    \item Compare the student's solution with the correct solution provided as a reference.
    \item Check if the student has correctly performed all key steps of the solution.
    \item If the student used a different approach, evaluate its correctness and compliance with the criteria.
    \item Problems with parameters allow for various solution methods: algebraic, geometric, functional.
    \item Pay \textbf{SPECIAL ATTENTION} to the correctness of handling boundary points and the completeness of considering all cases.
    \item When assessing for 3 points: check that the difference from the correct answer is \textbf{ONLY} in the boundary points, not in the main intervals.
    \item For a geometric approach: check the correctness of the interpretation and the justification of all geometric statements.
\end{itemize}

\subsubsection*{IMPORTANT: Instructions for working with the correct solution and the student's solution}
You are provided with:
\begin{enumerate}
    \item The correct solution to the task - use it as a reference for comparison.
    \item The student's solution - this is what you must evaluate.
\end{enumerate}

During the analysis:
\begin{itemize}
    \item Compare each step of the student's solution with the corresponding step of the correct solution.
    \item Note all deviations and errors.
    \item Check if the intermediate and final results match.
    \item Pay special attention to the handling of boundary points and the completeness of considering all cases.
\end{itemize}

\subsubsection*{Instructions for checking the solution of a problem with a parameter}
\begin{enumerate}
    \item Check the solution method:
    \begin{itemize}
        \item Correctness of the chosen approach (algebraic, geometric, functional).
        \item Correctness of applying formulas and theorems.
        \item Completeness of considering all cases.
    \end{itemize}
    \item Check mathematical correctness:
    \begin{itemize}
        \item Correctness of algebraic transformations.
        \item Correctness of working with inequalities.
        \item Correctness of finding the domain of permissible values (ODZ).
    \end{itemize}
    \item Check the handling of boundary points:
    \begin{itemize}
        \item Correctness of determining the boundary points.
        \item Correctness of including/excluding boundary points in the answer.
    \end{itemize}
    \item For a geometric approach, check:
    \begin{itemize}
        \item Correctness of the geometric interpretation of the conditions.
        \item Justification of all geometric statements.
        \item Completeness of the analysis of all possible relative positions of geometric objects.
    \end{itemize}
\end{enumerate}

\subsubsection*{CRITICALLY IMPORTANT: Immediately compare the student's answers with the correct ones!}
\begin{itemize}
    \item \textbf{FIRST AND FOREMOST}, check if the student's answer matches the correct answer.
    \item If the student's answer is \textbf{INCORRECT}, this \textbf{MUST} be taken into account in the assessment.
    \item Even if all transformations are performed correctly, but the answer is wrong due to a non-computational error - this must affect the score.
    \item Do not forget to note all discrepancies between the student's answer and the correct answer.
\end{itemize}

\subsubsection*{IMPORTANT: Distinguish between computational and conceptual errors}
\begin{itemize}
    \item \textbf{Computational errors:} errors in arithmetic operations, simplifying expressions, calculating values.
    \item \textbf{Conceptual errors:} incorrect application of formulas, wrong solution method, errors in understanding the properties of the parameter.
\end{itemize}
If a student made only computational errors, but the solution method is correct - this may correspond to the 2-point criterion.
If a student made conceptual errors - this usually corresponds to a lower score criterion.

\subsubsection*{Assessment Examples}

\paragraph{Example 1 (score: 4 points)}
The solution is complete and justified. All values of the parameter for which the system has exactly two solutions are found correctly. All cases are considered, and boundary points are analyzed correctly.

\paragraph{Example 2 (score: 3 points)}
All stages are present in the solution. Through correct reasoning, a set of parameter values is obtained that differs from the required set only by the exclusion of boundary points or the inclusion of points not belonging to the answer. For example, the answer obtained is $(-2,2)$ instead of $[-2,2]$ or $[-2,-1)\cup(-1,1)\cup(1,2]$ instead of $[-2,-1)\cup(-1,1)\cup[1,2]$.

\paragraph{Example 3 (score: 2 points)}
The solution is logical, all steps are present, but a computational error was made when solving the inequality, which led to an incorrect answer. Or, through correct reasoning, an interval of the set of parameter values was obtained, possibly including boundary points, but not all possible cases were considered.

\paragraph{Example 4 (score: 1 point)}
The roots of the equation are found in the solution, and the problem is correctly reduced to the investigation of these roots under the condition $x^2+ax+1 \geq 0$, but the further investigation is incorrect or absent.

\paragraph{Example 5 (score: 0 points)}
The solution contains fundamental errors in understanding the problem or in mathematical reasoning. For example, the roots of the original equation are found incorrectly, or the condition of the problem is misinterpreted.

\subsubsection*{IMPORTANT: Instructions for working with multiple images}
If the solution (correct or student's) is presented in several images:
\begin{enumerate}
    \item Consider all images as a single solution.
    \item Analyze them in sequential order.
    \item Take into account information from all images when assessing.
\end{enumerate}

\subsubsection*{Response Format}
Structure your response as follows:

\paragraph{Analysis of the solution}
[Detailed analysis indicating correct steps and errors]

\paragraph{Comparison with the correct solution}
[Comparison of the student's approach with the reference solution, indicating key differences]

\paragraph{Final score}
[Score: X points]
\vspace{1ex}

\textbf{IMPORTANT:} The "Final score" section \textbf{MUST} be formatted \textbf{EXACTLY} in this format:
"\verb|\paragraph{Final score}|" - heading.
"[Score: X points]" - where X is a number 0, 1, 2, 3, or 4.
\textbf{DO NOT CHANGE} this format and do not add any other text to it!
\vspace{1ex}

\paragraph{Justification of the score}
[Detailed justification in accordance with the criteria]

\paragraph{Recommendations (if applicable)}
[Recommendations for correcting errors]

\end{mdframed}

\vspace{1ex}

\subsubsection{Response from OpenAI o4-mini (Score: 2 - Correct)}
\vspace{1ex}
\begin{mdframed}[backgroundcolor=gray!5]
\small\ttfamily

\subsubsection*{Analysis of the solution}

\begin{enumerate}
    \item The student wrote down an equivalent system:
    \begin{itemize}
        \item a) the equation after squaring;
        \item b) the condition that the expression under the root is non\text{-}negative.
    \end{itemize}
    \item Found the roots of the original equation:
    $x = 0$, $x = 1\text{-}a$, $x = \text{-}1\text{-}a$.
    \item Analyzed the condition $x^2 + ax + 1 \geq 0$ for these values of x:
    \begin{itemize}
        \item for $x = 0$, the condition is always met;
        \item for $x = 1\text{-}a$, from the inequality, they obtained $a \leq 2$;
        \item for $x = \text{-}1\text{-}a$, they obtained $a \geq \text{-}2$.
    \end{itemize}
    \item Concluded that for three distinct roots, it is necessary that $\text{-}2 \leq a \leq 2$ and $a \neq 1$, $a \neq \text{-}1$.
    \item However, in the final answer, the student wrote
    $a \in (\text{-}1; 1) \cup (1; 2]$,
    thus completely omitting the interval $[\text{-}2; \text{-}1)$.
\end{enumerate}

\textbf{Error:} when combining the resulting intervals and excluding the boundary points, the student discarded the left part $[\text{-}2; \text{-}1)$ for a purely computational/descriptive reason. All key steps (finding roots, conditions on a, accounting for exclusions) were performed correctly, but the answer contains an inaccuracy.

\subsubsection*{Comparison with the correct solution}

\textbf{Reference answer:}
$a \in [\text{-}2; \text{-}1) \cup (\text{-}1; 1) \cup (1; 2]$.

The student obtained the same intermediate boundaries \text{-}2 and 2, and correctly excluded $a = \pm 1$, but missed the part $[\text{-}2; \text{-}1)$ in the final answer. Therefore, the method is correct, but a computational (descriptive) slip in combining the results led to an incomplete answer.

\subsubsection*{Final score}

[Score: 2 points]

\subsubsection*{Justification of the score}

According to the criteria:
\begin{itemize}
    \item All key steps of the solution (finding roots, domain of permissible values, conditions on the parameter, exclusion of boundary values) were performed correctly.
    \item The final answer is incorrect due to a computational/descriptive error when combining the intervals.
    \item The approach is mathematically correct, but the result is inaccurate.
\end{itemize}

This corresponds to the 2\text{-}point criterion: "An interval of the set of parameter values is obtained through correct reasoning, possibly including/excluding boundary points, but the overall answer is incorrect due to a computational error."

\subsubsection*{Recommendations}

\begin{itemize}
    \item Be more careful when combining intervals and check that no sections have been dropped.
    \item After obtaining the conditions on the parameter, explicitly write down all three 'pieces' and only then combine them, excluding the single points.
\end{itemize}

\end{mdframed}
\vspace{1ex}

\subsubsection{Response from Qwen 2.5 VL 32B (Score: 4 - Overestimated)}
\vspace{1ex}
\begin{mdframed}[backgroundcolor=gray!5]
\small\ttfamily

\subsubsection*{Analysis of the solution}

\paragraph{Step 1: Forming the system of equations}
The student correctly understands that for the initial equation $\sqrt{3x^2 + 2ax + 1} = x^2 + ax + 1$, two conditions must be met:
\begin{enumerate}
    \item $x^2 + ax + 1 \geq 0$ (since the square root is defined only for non\text{-}negative values).
    \item $3x^2 + 2ax + 1 = (x^2 + ax + 1)^2$ (the main equation).
\end{enumerate}

This approach is correct, and the student correctly transforms the equation:
$$
3x^2 + 2ax + 1 = x^4 + 2ax^3 + (a^2 + 2)x^2 + 2ax + 1.
$$
Then, they combine like terms and obtain:
$$
x^4 + 2ax^3 + (a^2 \text{-} 1)x^2 = 0.
$$
Next, the student correctly factorizes this equation:
$$
x^2(x^2 + 2ax + (a^2 \text{-} 1)) = 0.
$$
From this, they find the roots:
$$
x = 0 \quad \text{and} \quad x^2 + 2ax + (a^2 \text{-} 1) = 0.
$$
For the quadratic equation $x^2 + 2ax + (a^2 \text{-} 1) = 0$, the student finds the roots:
$$
x = \text{-}a + 1 \quad \text{and} \quad x = \text{-}a \text{-} 1.
$$
Thus, the student correctly finds three roots: $x = 0$, $x = \text{-}a + 1$, and $x = \text{-}a \text{-} 1$.

\paragraph{Step 2: Analysis of the conditions for distinct roots}
The student recognizes that for the equation to have three distinct roots, these roots must be different. They check the cases where the roots coincide:
\begin{enumerate}
    \item $x = 0$ and $x = \text{-}a + 1$: This is possible if $\text{-}a + 1 = 0 \Rightarrow a = 1$.
    \item $x = 0$ and $x = \text{-}a \text{-} 1$: This is possible if $\text{-}a \text{-} 1 = 0 \Rightarrow a = \text{-}1$.
    \item $x = \text{-}a + 1$ and $x = \text{-}a \text{-} 1$: This is possible if $\text{-}a + 1 = \text{-}a \text{-} 1$, but this is impossible as it leads to a contradiction.
\end{enumerate}

\paragraph{Step 3: Verifying the condition $x^2 + ax + 1 \geq 0$}
The student checks the condition $x^2 + ax + 1 \geq 0$ for each root:
\begin{enumerate}
    \item For $x = 0$: $x^2 + ax + 1 = 1 \geq 0$ is always met.
    \item For $x = \text{-}a + 1$: $x^2 + ax + 1 = (\text{-}a + 1)^2 + a(\text{-}a + 1) + 1 = a^2 \text{-} 2a + 1 \text{-} a^2 + a + 1 = \text{-}a + 2$. This is non\text{-}negative if $\text{-}a + 2 \geq 0 \Rightarrow a \leq 2$.
    \item For $x = \text{-}a \text{-} 1$: $x^2 + ax + 1 = (\text{-}a \text{-} 1)^2 + a(\text{-}a \text{-} 1) + 1 = a^2 + 2a + 1 \text{-} a^2 \text{-} a + 1 = a + 2$. This is non\text{-}negative if $a + 2 \geq 0 \Rightarrow a \geq \text{-}2$.
\end{enumerate}

\paragraph{Step 4: Combining the conditions}
The student correctly combines all the obtained conditions:
\begin{itemize}
    \item $a \neq 1$ and $a \neq \text{-}1$ (for distinct roots).
    \item $a \leq 2$ and $a \geq \text{-}2$ (to satisfy the condition $x^2 + ax + 1 \geq 0$).
\end{itemize}

Thus, the student obtains:
$$
a \in [\text{-}2, \text{-}1) \cup (\text{-}1, 1) \cup (1, 2].
$$

\subsubsection*{Comparison with the correct solution}

The correct solution and the student's solution completely coincide in the answer:
$$
a \in [\text{-}2, \text{-}1) \cup (\text{-}1, 1) \cup (1, 2].
$$
The student correctly performed all the key steps:
\begin{enumerate}
    \item Transforming the original equation.
    \item Finding the roots.
    \item Analyzing the conditions for distinct roots.
    \item Verifying the condition $x^2 + ax + 1 \geq 0$.
    \item Combining all the obtained conditions.
\end{enumerate}

\subsubsection*{Final score}
[Score: 4 points]

\subsubsection*{Justification of the score}
The student justifiably obtained the correct answer by correctly performing all the key steps of the solution. All intermediate results are correct, the boundary points are handled correctly, and all possible cases are considered. The solution method fully complies with the mathematical requirements of the problem, and the answer completely matches the correct solution.

\subsubsection*{Recommendations}
The student's solution contains no errors and requires no corrections. However, the student could improve the presentation by making it more detailed and structured, so that readers can better follow their reasoning. For example, they could explain in more detail why the roots coincide when $a = 1$ or $a = \text{-}1$, and why these values are excluded from the answer.

\end{mdframed}
\vspace{1ex}

\end{document}